%% file: main.tex
\documentclass[11pt,a4paper]{article}
\usepackage[hyperref]{conll-2019}
\usepackage{times}
\usepackage{latexsym}

\usepackage{enumitem}
\newlist{compactitem}{itemize}{3} % 3 is max-depth
\setlist[compactitem]{label=\textbullet, nosep}

\usepackage{hyperref}

\usepackage{cleveref}
\crefformat{section}{\S#2#1#3} % see manual of cleveref, section 8.2.1
\crefformat{subsection}{\S#2#1#3}
\crefformat{subsubsection}{\S#2#1#3}

\usepackage{url}
\usepackage{comment}
\usepackage[flushleft]{threeparttable}
\usepackage{graphicx}
\usepackage{amsmath}
\usepackage{subfig}
\usepackage{tabularx}
\usepackage{multirow}
\usepackage{algorithm}
\usepackage{todonotes}
\usepackage[noend]{algpseudocode}
\usepackage[hang,flushmargin]{footmisc}

\usepackage{amssymb}
\DeclareMathOperator{\E}{\mathbb{E}}

\usepackage{xcolor} 

\definecolor{MyColor}{RGB}{50, 100, 250}
\definecolor{Orange}{RGB}{244, 101, 66}
\definecolor{Green}{RGB}{255, 0, 0}

\title{Cross-lingual Dependency Parsing with Unlabeled Auxiliary Languages}

\aclfinalcopy % Uncomment this line for the final submission

\author{
Wasi Uddin Ahmad$^1$, Zhisong Zhang$^2$, Xuezhe Ma$^2$\\
\texttt{wasiahmad@cs.ucla.edu,\{zhisongz,xuezhem\}@cs.cmu.edu}\\
\textbf{Kai-Wei Chang}$^1$, \textbf{Nanyun Peng}$^3$\\
\texttt{kwchang@cs.ucla.edu,npeng@isi.edu} \\ [7pt]
$^1$University of California, Los Angeles, $^2$Carnegie Mellon University \\ $^3$University of Southern California\\
}

\begin{document}

\setlength{\abovedisplayskip}{5pt}
\setlength{\belowdisplayskip}{5pt}
\graphicspath{{images2/}}

\maketitle

\begin{abstract} 
% \NoteNPIL{suggested title: ``Finding Your Best Mates/Friends: Zero-shot Cross-lingual Dependency Parsing with Unlabeled Auxiliary Languages'' or ``Your Friends Know You More Than Yourself: Cross-lingual Dependency Parsing with Unlabeled Auxiliary Languages''. more brain storming are welcome!}

Cross-lingual transfer learning has become an important weapon to battle the unavailability of annotated resources for low-resource languages. 
One of the fundamental techniques to transfer across languages is learning \emph{language-agnostic} representations, in the form of word embeddings or contextual encodings.
In this work, we propose to leverage unannotated sentences from auxiliary languages to help learning language-agnostic representations. 
Specifically, we explore adversarial training for learning contextual encoders that produce invariant representations across languages to facilitate cross-lingual transfer.
We conduct experiments on cross-lingual dependency parsing where we train a dependency parser on a source language and transfer it to a wide range of target languages. 
Experiments on 28 target languages demonstrate that adversarial training significantly improves the overall transfer performances under several different settings. 
We conduct a careful analysis to evaluate the language-agnostic representations resulted from adversarial training. 

\end{abstract}

\input{introduction2}

\input{method}
\input{experiment}

\input{relwork}

\input{conclusion}

\subsubsection*{Acknowledgments}
We thank the anonymous reviewers for their helpful feedback.
This work was supported in part by National Science Foundation Grant IIS-1760523.

\bibliography{emnlp2019}
\bibliographystyle{acl_natbib}

\end{document}

%% file: introduction2.tex
\section{Introduction}
% \textbf{Paragraph theme:} We will first talk about cross-lingual NLP applications and emphasize on learning language agnostic representations to improve cross-lingual performances.

% In natural language processing (NLP), many state-of-the-art models are based on supervised learning approaches that rely on the availability of a copious amount of training data. 
% For resource-rich languages, ex., English, Chinese such labeled data is indeed available but for the majority of languages, however, only a limited amount of annotations exists.

% \notezs{(1) (why it is substantial) intro to cross-lingual transfer and also intro parsing here, is it too early?}

Cross-lingual transfer, where a model learned from one language is transferred to another, has become an important technique to improve the quality and coverage of natural language processing (NLP) tools for languages in the world. 
%Cross-lingual transfer, where knowledge learned from a resource-rich language is transferred to a low resource language, 
% Cross-lingual transfer learning has been widely studied in many NLP applications including POS tagging \cite{kim2017cross}, dependency parsing \cite{ma2014unsupervised}, named entity recognition \cite{xie2018neural}, entity linking \cite{sil2018neural}, coreference resolution \cite{kundu2018neural}, text classification \cite{xu-yang-2017-cross}, question answering \cite{joty2017cross} and noteworthy improvements are achieved on low resource language applications.
This technique has been widely applied in many applications,
% \NoteMA{this sentence seems to suffer grammartical errors. Should it be change to "Cross-lingual transfer, where knowledge learned from a ..., has been studied in many NLP applications ..."}
% ~\NoteNP{Can you name the applications and cite the papers for different applications separately?} 
% \cite{kim2017cross, ma2014unsupervised, xie2018neural, sil2018neural, kundu2018neural, xu-yang-2017-cross, joty2017cross}
including part-of-speech (POS) tagging \cite{kim2017cross}, dependency parsing \cite{ma2014unsupervised}, named entity recognition \cite{xie2018neural}, entity linking \cite{sil2018neural}, coreference resolution \cite{kundu2018neural}, and question answering \cite{joty2017cross}.
Noteworthy improvements are achieved on low resource language applications due to cross-lingual transfer learning. 
% In this paper, we study cross-lingual transfer for dependency parsing, since it is one of the core NLP tasks and the development of Universal Dependencies \cite{nivre2016universal} provides consistent annotations across languages, allowing us to investigate transfer learning in a wide range of languages.

% \notezs{(2) (motivation) current approach and their limitation and bring in the motivation for learning language-agnostic representation.}
%In neural transfer learning for NLP, one of the common practices is to transfer a neural network component across domains or languages.
%In cross-lingual transfer, such a component is trained to learn language representations so that it can benefit the low-resource language applications.
%In the core of deep neural models for NLP, a component is designated to transform the input text sequences into hidden vectors that consider the contextual and structured information of the input, known as the \emph{encoder}.
% In the core of neural models for NLP, a component is designated to transform the input text sequences into hidden vectors that consider the contextual and structured information of the input, known as the \emph{encoder}.
% One of the main underneath challenges in building such cross-lingual models is training the encoders to learn cross-lingual regularities. 

In this paper, we study cross-lingual transfer for dependency parsing. A dependency parser consists of (1) an encoder that transforms an input text sequence into latent representations and (2) a decoding algorithm that generates the corresponding parse tree. 
In cross-lingual transfer, most recent approaches assume that the inputs from different languages are aligned into the same embedding space via multilingual word embeddings or multilingual contextualized word vectors, such that the parser trained on a source language can be transferred to target languages. 
% \notezs{I would suggest us to be careful to use the saying of `semantic space`.}
%KW: It doesn't make sense to require the parser only capture semantic information as  parsing is a syntactic-oriented task. 
However, when training a parser on the source language, the encoder not only learns to embed a sentence but it also carries language-specific properties, such as word order typology. Therefore, the parser suffers when it is transferred to a language with different language properties. 
%\emph{semantic} information, but also \emph{syntactic} and \emph{structural} information of a language, such as word order typology. 
%Therefore, the encoder trained on the source language may carries 
%can overfit to their specific structured patterns, which should not be transferred to other languages.
% This motivates our work of training encoders to learn shared structural patterns across languages and make them more robust when transferring to target languages.
Motivated by this, we study how to train an encoder for generating language-agnostic representations that can be transferred across a wide variety of languages.

% \notezs{(3) (our proposed approach) To build the agnostic representation, we consider using an auxiliary task of predicting language ID with adversarial training}

% We propose to take advantages of \emph{unlabeled} sentences of one or more auxiliary languages apart from the source language to train the encoder such that it becomes less language-specific and facilitates cross-lingual transfer.
We propose to utilize \emph{unlabeled} sentences of one or more auxiliary languages to train an encoder that learns language-agnostic contextual representations of sentences to facilitate cross-lingual transfer.
%The encoder is a part of a deep neural model that is primarily built for an NLP task and trained on the source language.
%Our goal is to train the encoder using source and auxiliary language sentences such that it learns to capture generic language features  to facilitate cross-lingual transfer.
% \notezs{maybe we should be consistent on the saying of `unlabeled` or `unannotated`?} - handled (wasi)
% We call those additional languages as \emph{auxiliary languages}.
To utilize the unlabeled auxiliary language corpora, we adopt adversarial training \cite{goodfellow2014generative} of the encoder
% to generate representation that is fed to a classifier to predict the language identity given an input sentence.
and a classifier that predicts the language identity of an input sentence from its encoded representation produced by the encoder.
% contextual representations of an input sentence 
The adversarial training encourages the encoder to produce language invariant representations such that the language classifier fails to predict the correct language identity.
% The goal of the adversarial training is to train the encoder to produce language invariant representations such that the language classifier fails to predict the correct language identity.
As the encoder is jointly trained with a loss for the primary task on the source language and adversarial loss on all languages, we hypothesize that it will learn to capture task-specific features as well as generic structural patterns applicable to many languages, and thus have better transferrability.
To verify the proposed approach, we conduct experiments on neural dependency parsers trained on English (source language) and directly transfer them to 28 target languages,
% from the Universal Dependencies \cite{ud22} dataset
with or without the assistance of unlabeled data from auxiliary languages.
We chose dependency parsing as the primary task since it is one of the core NLP applications and the development of Universal Dependencies \cite{nivre2016universal} provides consistent annotations across languages, allowing us to investigate transfer learning in a wide range of languages.
% \notezs{Maybe split this sentence, too complex?}
Thorough experiments and analyses are conducted to address the following research questions:

\vspace{5pt}
\begin{compactitem}
    \item Does encoder trained with adversarial training generate language-agnostic  representations? 
    \item Does language-agnostic representations improve cross-language transfer?
\end{compactitem}
\vspace{5pt}

Experimental results show that the proposed approach consistently outperform a strong baseline parser \cite{ahmad2019}, with a significant margin in two family of languages. 
% ~\NoteNP{Do we discussed about the language family issue? If not, we probably don't want to say this in the intro. If we do, then for table 2, I advocate to change the ``Dist. to English'' column to be ``Language Family'' and arrange the languages by family.}
In addition, we conduct experiments to consolidate our findings with different types of input representations and encoders.
Our experiment code is publicly available to facilitate future research.\footnote{https://github.com/wasiahmad/cross\_lingual\_parsing}

% Our work, therefore, provides a strong basis for adversarial training to improve cross-lingual transfer for different NLP tasks.~\NoteNP{I won't put such a strong ending in the intro.}

% Our work therefore provides a strong basis to utilize adversarial training by using unlabeled language corpora to train the neural encoders to learn generic language representations that can improve cross-lingual transfer across a wide range of languages for different NLP tasks.

% Furthermore, our analyses reveal the following insights on the adversarial training for cross-lingual transfer parsing.

% \vspace{2pt}
% \begin{compactitem}
%     \item For order-free models, if we pick languages distant from the source language as auxiliary language, adversarial training results in larger improvements and vice versa.
%     \item Languages having smaller average distance to all the target languages are better candidate as auxiliary languages.
%     \item Even if we have richer multilingual word representations, language adaptation using auxiliary languages can still help the encoders to learn language independent representations.
% \end{compactitem}
% \vspace{2pt}

%% file: method.tex
%\section{Unsupervised Language Adaptation for Dependency Parsing}
\section{Training Language-agnostic Encoders}
% \NoteNP{I won't call it unsupervised language adaptation. Adaptation gives people an impression that we are adapting to \emph{each} target languages.}

\begin{figure}[t]
\centering
\includegraphics[width=1.0\linewidth]{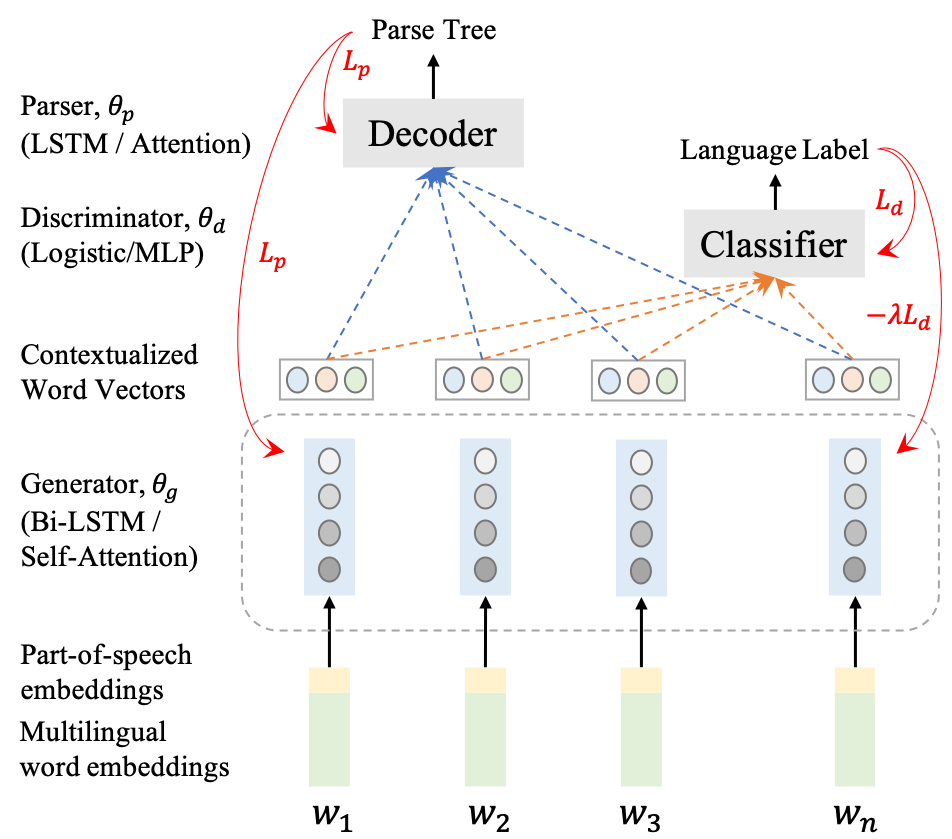}
\caption{An overview of our experimental model consists of three basic components: (1) Encoder, (2) (Parsing) Decoder, and (3) (Language) Classifier.
We also show how parsing and adversarial losses ($L_p$ and $L_d$) are back propagated for parameter updates.
}
\label{fig:model}
% \vspace{-1em}
\end{figure}

%KW: the following two sentences repeats intro.
%Our primary goal is to improve cross-lingual transfer by inducing a language invariant structural encoder, such that it does not overfit to the structural information (e.g., word order typology) of the source language and thus generalizes better to a wide variety of target languages. %applying adversarial training to the representation learning component, known as the \emph{encoder} in the dependency parsing model.
%To achieve this goal, we propose to leverage unlabeled sentences in one or more languages other than the source language to train an encoder to become language-agnostic.
We train the encoder of a dependency parser in an \emph{adversarial} fashion to guide it to avoid capturing language-specific information. 
In particular, we introduce a language identification task where a classifier predicts the language identity (id) of an input sentence from its encoded representation. 
Then the encoder is trained such that the classifier fails to predict the language id while the parser decoder predicts the parse tree accurately from the encoded representation.
%We train the encoder in an \emph{adversarial} fashion to guide it to avoid capturing language-specific information such that the language id classifier fails. With the adversarial training, 
We hypothesize that such an encoder would have better cross-lingual transferability.
The overall architecture of our model is illustrated in Figure \ref{fig:model}.
In the following, we present the details of the model and training method.

\subsection{Architecture}
\label{sec:par}
% \textbf{Theme:} We will briefly introduce the two architectures, SelfAttn-Graph and RNN-StackPtr parsers which we will use in our experiments.
% \notezs{I keep the model introduction brief since they are not our focus, please check whether sth are missing.}

Our model consists of three basic components, (1) a general encoder, (2) a decoder for parsing, and (3) a classifier for language identification.
The encoder learns to generate contextualized representations for the input sentence (a word sequence) which are fed to the decoder and the classifier to predict the dependency structure and the language identity (id) of that sentence.
% While the encoder collaborates with the decoder to predict the dependency parsing tree, it acts adversarially so that the classifier fails to predict the language identity (id) given the input sequences.

The encoder and the decoder jointly form the parsing model and we consider two alternatives\footnote{\citet{ahmad2019} studied order-sensitive and order-free models and their performances in cross-lingual transfer. In this work, we adopt two typical ones and study the effects of adversarial training on them.} from \cite{ahmad2019}: ``SelfAtt-Graph'' and ``RNN-Stack''.
The ``SelfAtt-Graph'' parser consists of a modified self-attentional encoder \cite{shaw2018self} and a graph-based deep bi-affine decoder \cite{dozat2017biaffiine}, while the ``RNN-Stack'' parser is composed of a Recurrent Neural Network (RNN) based encoder and a stack-pointer decoder \cite{ma2018stack}.

We stack a classifier (a linear classifier or a multi-layer Perceptron (MLP)) on top of the encoder to perform the language identification task. 
The identification task can be framed as either a word- or sentence-level classification task.
For the sentence-level classification, we apply average pooling\footnote{We also experimented with max-pooling and weighted pooling but average pooling resulted in stable performance.} on the contextual word representations generated by the encoder to form a fixed-length representation of the input sequence, which is fed to the classifier. For the word-level classification, we perform language classification for each token individually.
% \notezs{Have we tried other poolings such as max-pool? If so, maybe adding a footnote?}

In this work, following the terminology in adversarial learning literature, we interchangeably call  the encoder as
% In our later adversarial treating, we will regard the encoder as 
the generator, G and the classifier as the discriminator, D.

% In our later adversarial treating, we will regard the encoder as the generator G, for generating representations for the input sentences and the classifier as the discriminator D, for distinguishing different language sentences.

% \paragraph{Classifier}
% \smallskip
% \noindent\textbf{$\bullet$ Classifier}
% We stack a classifier (a linear classifier or a multi-layer Perceptron (MLP)) on top of the encoder so that the classifier predicts the language identity (id) of the input sequences from the contextualized representations. 
% We can frame the language prediction task as either a word- or sentence-level classification task.
% For sentence-level classification, we apply average pooling on the contextual word representation to form a fixed length vector representation of the input sequence, which is fed to the classifier.

% Based on the contextual representations generated by the encoder, we stack a language classifier (a linear classifier or a multi-layer Perceptron (MLP)) on it to predict the language identity (id) of the input sequence. 
% We can frame the language prediction task as either a word- or sentence-level classification task.
% For sentence-level classification, we apply average pooling on the contextual word representation to form a fixed length vector representation of the input sequence, which is fed to the classifier.
% We explore two alternatives, a linear classifier and a Multi-layer Perceptron (MLP) as the language classifier. 
% This classifier plays the role of the discriminator D in our later adversarial training.

\begin{algorithm}[t]
% \small
\caption{Training procedure.}
\label{alg:gan}
Parameters to be trained: Encoder ($\theta_g$), Decoder  ($\theta_p$), and Classifier ($\theta_d$) \\
$X^a$ = Annotated source language data \\
$X^b$ = Unlabeled auxiliary language data \\
$I$ = Number of warm-up iterations \\
$k$ = Number of learning steps for the discriminator ($D$) at each iteration \\
% $\mathcal{L}_p$ = parsing loss; $\mathcal{L}_d$ = classification loss; $\lambda$ = Coefficient of $\mathcal{L}_d$ \\
$\lambda$ = Coefficient of $\mathcal{L}_d$ \\
$\alpha_1$, $\alpha_1$ = learning rate; $B$ = Batch size
\begin{algorithmic}[1]
\Require
\For{$j = 0,\cdots,I$}
    \State Update $\theta_g := \theta_g - \alpha_1 \nabla_{\theta_g} \mathcal{L}_p$ \State Update $\theta_p := \theta_p - \alpha_1 \nabla_{\theta_p} \mathcal{L}_p$
    % $O_t$ as shown in Eq~\ref{eq:parser_objective}.
\EndFor
\For{$j = I,\cdots,num\_iter$}
    \For{$k$ steps}
        \State  $(x_{a}^i)^{B/2}_{i=1}\gets $ Sample a batch from $X^a$
        \State  $(x_{b}^i)^{B/2}_{i=1}\gets $ Sample a batch from $X^b$
        \State Update $\theta_d := \theta_d - \alpha_2 \nabla_{\theta_d} \mathcal{L}_d$
    \EndFor
    \State Total loss $\mathcal{L} := \mathcal{L}_p - \lambda \mathcal{L}_d$
    \State Update $\theta_g := \theta_g - \alpha_1 \nabla_{\theta_g} \mathcal{L}$ 
    % and \\ $\theta_p := \theta_p - \alpha_1 \nabla_{\theta_p} \mathcal{L}$
    \State Update $\theta_p := \theta_p - \alpha_1 \nabla_{\theta_p} \mathcal{L}$
\EndFor
\end{algorithmic}
\end{algorithm}

\subsection{Training}

Algorithm \ref{alg:gan} describes the training procedure. We have two types of loss functions: $\mathcal{L}_p$ for the parsing task and $\mathcal{L}_d$ for the language identification task. For the former, we update the encoder and the decoder as in the regular training of a parser. For the latter, we adopt adversarial training to update the encoder and the classifier. We present the detailed training schemes in the following.

\subsubsection{Parsing} 
To train the parser, we adopt both cross-entropy objectives for these two types of parsers as in \cite{dozat2017biaffiine,ma2018stack}. The encoder and the decoder are jointly trained to optimize the probability of the dependency trees ($y$) given sentences ($x$): 
$$\mathcal{L}_p = -\log p(y|x).$$ 
The probability of a tree can be further factorized into the products of the probabilities of each token's ($m$) head decision ($h(m)$) for the graph-based parser, or the probabilities of each transition step decision ($t_i$) for the transition-based parser:
\begin{align*}
    \text{Graph:~~~~} \mathcal{L}_p &= -\sum\nolimits_{m}\log p(h(m)|x, m), \\
    \text{Transition:~~~~} \mathcal{L}_p &= -\sum\nolimits_{i}\log p(t_i|x, t_{<i}).
\end{align*}

\subsubsection{Language Identification}
Our objective is to train the contextual encoder in a dependency parsing model such that it encodes language specific features as little as possible, which may help cross-lingual transfer.
To achieve our goal, we utilize adversarial training by employing unlabeled auxiliary language corpora.
% We study several different techniques of adversarial training and briefly describe them as follows.

% \notezs{I describe GAN as ``basic setup'' and other two  as ``alternatives''. The alternatives are described with shorter descriptions. Please check whether this is fine. Another thing: I'm not sure how GR is different than GAN: the only diff is one output vs. two outputs for the classification? which I think is similar to GAN and maybe we don't need to give detailed formula.}

\paragraph{Setup}
% \smallskip
% \noindent\textbf{$\bullet$ Training with GAN}
We adopt the basic generative adversarial network (GAN) for the adversarial training.
We assume that $X^a$ and $X^b$ be the corpora of the source and auxiliary language sentences, respectively.
The discriminator acts as a binary classifier and is adopted to distinguish the source and auxiliary languages.
For the training of the discriminator, weights are updated according to the original classification loss:
% \begin{equation*}
\begin{align*}
\label{eq:gan}
    \mathcal{L}_d &= \E_{x\sim X^a} [\log\ D(G(x)] + \\  
    & \E_{x\sim X^b} [\log\ (1 - D(G(x))].
\end{align*}
% \end{equation*}

% To improve the stability of GAN based learning, \citet{arjovsky2017wasserstein} proposed an alternative approach called Wasserstein GAN (WGAN) which we also study in this work.
% According to the WGAN principle, we use the following loss function to update the discriminator weights.
% \begin{equation*}
% \label{eq:wgan}
%     \begin{aligned}
%         \mathcal{L}_d(\theta_d) = \E_{x\sim X^a} [D(G(x)] -  \E_{x\sim X^b} [D(G(x)]
%     \end{aligned}
% \end{equation*}

For the training of dependency parsing, the generator, $G$ collaborates with the parser but acts as an adversary with respect to the discriminator.
Therefore, the generator weights ($\theta_g$) are updated by minimizing the loss function, 
$$\mathcal{L} = \mathcal{L}_p - \lambda \mathcal{L}_d,$$ where $\lambda$ is used to scale the discriminator loss ($\mathcal{L}_d$).
In this way, the generator is guided to build language-agnostic representations in order to fool the discriminator while being helpful for the parsing task. Meanwhile, the parser can be guided to rely more on the language-agnostic features.

% \paragraph{Training with WGAN}
% \smallskip
% \noindent\textbf{$\bullet$ Training with WGAN}
% In WGAN \cite{arjovsky2017wasserstein} setup, we use the following loss function to update the discriminator weights.
% \begin{equation*}
% \label{eq:wgan}
%     \begin{aligned}
%         \mathcal{L}_d(\theta_d) = \E_{x\sim X^a} [D(G(x)] -  \E_{x\sim X^b} [D(G(x)]
%     \end{aligned}
% \end{equation*}

\paragraph{Alternatives} We also consider two alternative techniques for the adversarial training: Gradient Reversal (GR) \cite{ganin2016domain} and Wasserstein GAN (WGAN) \cite{arjovsky2017wasserstein}.
% \NoteMA{citation missed for GR and WGAN.}
% \paragraph{Training with Gradient Reversal (GR)}
% \smallskip
% \noindent\textbf{$\bullet$ Training with Gradient Reversal (GR)}
As opposed to GAN based training, in GR setup, the discriminator acts as a multiclass classifier that predicts language identity of the input sentence, and we use multi-class cross-entropy loss.
% \begin{equation*}
% \label{eq:gr}
%     \mathcal{L}_d = \E_{x\sim X^a \cup X^b} - \log D(G(x))[c]
%     % \mathcal{L}_d = - \sum_{c=1}^N y_c \log D(G(x))_c
% \end{equation*}
% where $x$ and $x'$ indicates the input word representations and the contextualized word vectors respectively, $N$ is the total number of source and auxiliary languages and $y_c$ is a binary indicator (0 or 1) if class label $c$ is the correct classification for the input sentence.
We also study Wasserstein GAN (WGAN), which is proposed by \citet{arjovsky2017wasserstein} to improve the stability of GAN based learning. Its loss function is shown as follows.
\begin{equation*}
\label{eq:wgan}
    \begin{aligned}
        \mathcal{L}_d = \E_{x\sim X^a} [D(G(x)] -  \E_{x\sim X^b} [D(G(x)],
    \end{aligned}
\end{equation*}
here, the annotations are similar to those in the GAN setting.

% In GR based parser training, we update the generator weights as: $\theta_g := \theta_g + \nabla_{\theta_g} \mathcal{L}_p - \lambda \nabla_{\theta_g} \mathcal{L}_d$ where $\lambda$ is used to scale the discriminator gradient.

% In GR based parser training, we reverse the gradient from the discriminator to update the generator weights as: $\theta_g := \theta_g + \nabla_{\theta_g} \mathcal{L}_p - \lambda \nabla_{\theta_g} \mathcal{L}_d$
% where $\lambda$ is used to scale the discriminator gradient.

% Algorithm \ref{alg:gan} describes the procedure to train the three components (Encoder, Decoder, and Classifier) of our model.

%% file: experiment.tex
\section{Experiments and Analysis}
\label{sec:ea}

% In this section, we discuss our experiments and analysis on cross-lingual dependency parsing transfer from a wide variety of perspectives.
% We conduct extensive evaluations showing the advantages of adversarial training.
In this section, we discuss our experiments and analysis on cross-lingual dependency parsing transfer from a variety of perspectives and show the advantages of adversarial training.

\input{table/table1.tex}

\paragraph{Settings.}
In our experiments, we study single-source parsing transfer, where a parsing model is trained on one source language and directly applied to the target languages.
We conduct experiments on the Universal Dependencies (UD) Treebanks (v2.2) \cite{ud22} using 29 languages, as shown in Table \ref{tab:langs}.
We use the publicly available implementation\footnote{https://github.com/uclanlp/CrossLingualDepParser} of the ``SelfAtt-Graph'' and ``RNN-Stack'' parsers.\footnote{We adopt the same hyper-parameters, experiment settings and evaluation metrics as those in \cite{ahmad2019}.} \citet{ahmad2019} show that the ``SelfAtt-Graph'' parser captures less language-specific information and performs better than the `RNN-Stack'' parser for distant target languages. Therefore, we use the ``SelfAtt-Graph'' parser in most of our experiments. Besides, the multilingual variant of BERT (mBERT) \cite{devlin2018bert} has shown to perform well in cross-lingual tasks \cite{wu2019beto} and outperform the models trained on multilingual word embeddings by a large margin.
%We take this as an inspiration and compare contextual encoders trained on top of mBERT and multilingual word embeddings.
Therefore, we consider conducting experiments with both multilingual word embeddings and mBERT.  
We use aligned multilingual word embeddings \cite{smith2017offline, bojanowski2017enriching} with $300$ dimensionss or contextualized word representations provided by multilingual BERT\footnote{https://github.com/huggingface/pytorch-transformers} \cite{devlin2018bert} with $768$ dimensions as the word representations. 
In addition, we use the Gold universal POS tags to form the input representations.\footnote{We concatenate the word and POS representations. In our future work, we will conduct transfer learning for both POS tagging and dependency parsing.}
% \notezs{Add a footnote to explain the strategy of Gold POS, maybe exploring other strategies for future work?}
We freeze the word representations during training to avoid the risk of disarranging the multilingual representation alignments.

\input{table/table2.tex}

% \begin{figure}[t]
% 	\centering
% 	\includegraphics[width=0.5\textwidth]{lang_tsne.pdf}
% 	%\vspace*{-4mm}
% 	\caption{t-SNE visualization \cite{maaten2008visualizing} of the languages by their word-ordering vectors.}
% 	\label{fig:lang_cluster}
% \end{figure}

We select six auxiliary languages\footnote{We want to cover languages from different families and with varying distances from the source language (English).} (French, Portuguese, Spanish, Russian, German, and Latin) for unsupervised language adaptation via adversarial training.
We tune the scaling parameter $\lambda$ in the range of $[0.1, 0.01, 0.001]$ on the source language validation set and report the test performance with the best value.
For gradient reversal (GR) and GAN based adversarial objectives, we use Adam \cite{kingma2014adam} to optimize the discriminator parameters, and for WGAN, we use RMSProp \cite{tieleman2012lecture}.
The learning rate is set to $0.001$ and $0.00005$ for Adam and RMSProp, respectively.
We train the parsing models for $400$ and $500$ epochs with multilingual BERT and multilingual word embeddings respectively.
We tune the parameter $I$ (as shown in Algorithm \ref{alg:gan}) in the range of $[50, 100, 150]$.

\paragraph{Language Test.} 
The goal of training the contextual encoder adversarially with unlabeled data from auxiliary languages is to encourage the encoder to capture more language-agnostic representations and less language-dependent features.
To test whether the contextual encoders retain language information after adversarial training, we train a multi-layer Perceptron (MLP) with softmax on top of the \emph{fixed} contextual encoders to perform a 7-way classification task.\footnote{With the source (English) and six auxiliary languages.}
If a contextual encoder performs better in the language test, it indicates that the encoder retains language specific information.

\subsection{Results and Analysis}

Table \ref{tab:res_test} presents the main transfer results of the ``SelfAtt-Graph'' parser when training on only English (en, baseline), English with French (en-fr), and English with Russian (en-ru). 
The results demonstrate that the adversarial training with the auxiliary language identification task benefits cross-lingual transfer with a small performance drop on the source language. When multi-lingual embedding is employed, the performance significantly improves, in terms of UAS of 0.48 and 0.61 over the 29 languages when French and Russian are used as the auxiliary language, respectively.
% \notezs{Maybe we should exclude English for the Average since we care more about transfer results? Has this been done?}
%performance gets improved from 
% The results demonstrate that both ``SelfAtt-Graph'' and ``RNN-Stack'' parsing models get benefits from the adversarial training with the auxiliary language identification task.
% \footnote{In most of our experiments, we use multilingual word embeddings as the word representations because the model training is faster comparing to using mBERT.}
When richer multilingual representation technique like mBERT is employed, adversarial training can still improve cross-lingual transfer performances (0.21 and 0.54 UAS over the 29 languages by using French and Russian, respectively).

Next, we apply adversarial training on the ``RNN-Stack'' parser and show the results in Table \ref{tab:res_rnnstack}. 
Similar to the ``SelfAtt-Graph''parser, the ``RNN-Stack'' parser resulted in significant improvements in cross-lingual transfer from  unsupervised language adaptation.
We discuss our detailed experimental analysis in the following.

\subsubsection{Impact of Adversarial Training}
\label{sec:impact_AT}

% \notezs{Table 7? The table is too far away.}
To understand the impact of different adversarial training types and objectives, we apply adversarial training on both word- and sentence-level with gradient reversal (GR), GAN, and WGAN objectives.
We provide the average cross-lingual transfer performances in Table \ref{table:at_abl} for different adversarial training setups.
Among the adversarial training objectives, we observe that in most cases, the GAN objective results in better performances than the GR and WGAN objectives.
% We present the average cross-lingual transfer performances for different AT objectives in Table \ref{table:at_abl}.
Our finding is in contrast to \citet{adel2018adversarial} where GR was reported to be the better objective.
% In GR, we reverse the gradient w.r.t the loss of the discriminator, and we argue that such minimax optimization may not provide an accurate signal to the encoders to learn language invariant features. 
% On the contrary, using GAN objective, the encoders get signal to capture features that make the source and auxiliary language distributions indistinguishable.
To further investigate, we perform the language test on the encoders trained via these two objectives. We find that the GR-based trained encoders perform consistently better than the GAN based ones on the language identification task, showing that via GAN-based training, the encoders become more language-agnostic.
In a comparison between GAN and WGAN, we notice that GAN-based training consistently performs better.

Comparing word- and sentence-level adversarial training, we observe that predicting language identity at the word-level is slightly more useful for the ``SelfAtt-Graph'' model, while the sentence-level adversarial training results in better performances for the ``RNN-Stack'' model. There is no clear dominant strategy.
% We suspect this is due to the order-free and order-sensitive nature of the respective models.
% For example, the contextual encoder in the ``RNN-Stack'' parser better encodes word order specific features, and it may help the discriminator to differentiate between languages. 
% As a result, the discriminator may fail to provide useful signals to the RNN encoder.
% This may be explained by the order-free and order-sensitive nature of the respective models: through word- and sentence-level training, the discriminator, can provide more useful signals to the self-attentional and RNN encoders respectively.

In addition, we study the effect of using a linear classifier or a multi-layer Perceptron (MLP) as the discriminator and find that the interaction between the encoder and the linear classifier resulted in improvements.\footnote{This is a known issue in GAN training as the discriminator becomes too strong, it fails to provide useful signals to the generator. In our case, MLP as the discriminator predicts the language labels with higher accuracy and thus fails.}
% In contrast, MLP as the discriminator predicts the language labels with higher accuracy, and fails to provide a useful signal to lead the encoder to build language agnostic features.

% \notezs{I suggest say our main focus first, which I guess is the impacts of aux-lang?, and leave the parts of model ablation (for example, adversarial methods/Linear-vs.-MLP) to a later subsection. For example, like 3.2.1 Main results and probing results, 3.2.2 impact of aux lang, 3.2.* vs. Multi-task, model ablation, mbert}

\input{table/table5.tex}

\input{table/table6.tex}

\input{table/table4.tex}

\input{table/table3.tex}

\input{table/table7.tex}

\subsubsection{Adversarial v.s. Multi-task Training}
\label{sec:adv_vs_mtl}
% Theme: Adversarial training utilizing auxiliary language data, we observe cross-lingual performance improvement. However, this results in a important question, does the improvements comes from adversarial training or only because the encoder observes more data? Instead of adversarial training, adding a motivational loss to the overall objective can be a baseline for our experiments. We want to show that adversarial training indeed helps to learn language agnostic representation where a motivational training would do the opposite. And in both of these settings, the amount of training data is same.
In section \ref{sec:impact_AT}, we study the effect of learning language-agnostic representation by using auxiliary language with adversarial training. 
An alternative way to leverage auxiliary language corpora is by encoding language-specific information in the representation via multi-task learning. 
% Interestingly, although this objective sounds contradiction to adversarial learning, it has a positive effect on the cross-lingual parsing, as the representations are learned with certain additional information from new (unlabeled) data. 
In the multi-task learning (MTL) setup, the model observes the same amount of data (both labeled and unlabeled) as the adversarially trained (AT) model.
The only difference between the MTL and AT models is that in the MTL models, the contextual encoders are encouraged to capture language-dependent features while in the AT models, they are trained to encode language-agnostic features.
% The only difference between the MTL and AT models is that in the MTL model, the contextual encoders collaborate with the language classifier instead of acting adversarially with the classifier as in the AT models.
% In essence, in multi-task learning, the contextual encoders are encouraged to capture language-dependent features.

The experiment results using multi-task learning in comparison with the adversarial training are presented in Table \ref{table:at_vs_mtl}.
Interestingly, although the MTL objective sounds contradiction to adversarial learning, it has a positive effect on the cross-lingual parsing, as the representations are learned with certain additional information from new (unlabeled) data. 
Using MTL, we sometimes observe improvements over the baseline parser, as indicated with the $\dagger$ sign, while the AT models consistently perform better than both the baseline and the MTL model (as shown in Columns 2--5 in Table \ref{table:at_vs_mtl}).
The comparisons on parsing performances do not reveal whether the contextual encoders learn to encode language-agnostic or dependent features.

Therefore, we perform language test with the MTL and AT (GAN based) encoders, and the results are shown in Table \ref{table:at_vs_mtl}, Columns 6--7.
The results indicate that the MTL encoders consistently perform better than the AT encoders, which verifies our hypothesis that adversarial training motivates the contextual encoders to 
% avoid language-dependent features and 
encode language-agnostic features.
% As we can see the language classifier predicts the language identity accurately when trained on top of the MTL encoders compared to the AT encoders. 
% This result confirms our hypothesis that via adversarial training, contextual encoders get motivated to capture language invariant features.

\subsubsection{Impact of Auxiliary Languages}
To analyze the effects of the auxiliary languages in cross-language transfer via adversarial training, we perform experiments by pairing up\footnote{We also conduct experiments on multiple languages as the auxiliary language. For GAN and WGAN-based training, we concatenate the corpora of multiple languages and treat them as one auxiliary language. In these set of experiments, we do not observe any apparent improvements.} the source language (English) with six different languages (spanning Germanic, Romance, Slavic, and Latin language families) as the auxiliary language. 
The average cross-lingual transfer performances are presented in Table \ref{table:lang_comp} and the results suggest that Russian (ru) and German (de) are better candidates for auxiliary languages.

We then dive deeper into the effects of auxiliary languages trying to understand whether auxiliary languages particularly benefit target languages that are closer to them\footnote{The language distances are computed based on word order characteristics as suggested in \citet{ahmad2019}.} or from the same family.
% We expected that an auxiliary language belonging to a language family would benefit other languages from the same family but we did not see any such pattern when we looked into the detailed results.
% Intuitively, if the target languages belong to the same language family as the auxiliary language, the cross-lingual transfer performance would be higher.
% Similarly, we expected that selecting a language as the auxiliary language, closer to all the target languages would perform better.
Intuitively, we would assume when the auxiliary language has a smaller average distance to all the target languages, the cross-lingual transfer performance would be better.
However, from the results in Table \ref{table:lang_comp}, we do not see such a pattern. For example, Portuguese (pt) has the smallest average distance to other languages among the auxiliary languages we tested, but it is not among the better auxiliary languages.

We further zoom in the cross-lingual transfer improvements for each language families as shown in Table \ref{tab:lang_family_test}.
% However, our experimental findings suggest that all these factors do not influence the cross-lingual transfer performance. 
We hypothesis that the auxiliary languages to be more helpful for the target languages in the same family. 
The experimental results moderately correlate with our expectation. Specifically, the Germanic family benefits the most from employing German (de) as the auxiliary language; similarly Slavic family with Russian (ru) as the auxiliary language (although German as the auxiliary language brings similar improvements). 
The Romance family is an exception because it benefits the least from using French (fr) as the auxiliary language. This may due to the fact that French is too closed to English, thus is less suitable to be used as an auxiliary language.

%% file: table/table1.tex
\begin{table}[t]
	\centering
	\resizebox{\linewidth}{!}{%
	\small
	\begin{tabular}{>{\centering\arraybackslash}p{1.8cm} | >{\centering\arraybackslash}p{4.75cm}}
		\hline
		Language Families & Languages \\
		\hline
		Afro-Asiatic & Arabic (ar), Hebrew (he)\\
		\hline
		Austronesian & Indonesian (id)\\
		\hline
		IE.Baltic & Latvian (lv)\\
		\hline
		IE.Germanic & Danish (da), Dutch (nl), English (en), German (de), Norwegian (no), Swedish (sv)\\
		\hline
		IE.Indic & Hindi (hi)\\
		\hline
		IE.Latin & Latin (la)\\
		\hline
		IE.Romance & Catalan (ca), French (fr), Italian (it), Portuguese (pt), Romanian (ro), Spanish (es)\\
		\hline
		IE.Slavic & Bulgarian (bg), Croatian (hr), Czech (cs), Polish (pl), Russian (ru), Slovak (sk), Slovenian (sl), Ukrainian (uk)\\
		\hline
		Korean & Korean (ko)\\
		\hline
		Uralic & Estonian (et), Finnish (fi)\\
		\hline
	\end{tabular}
}
\caption{
\label{tab:langs} 
The selected 29 languages for experiments from UD v2.2 \cite{ud22}.
}
% \caption{\label{tab:langs} The selected 29 languages grouped based on language families (adapted from \citet{ahmad2019}). ``IE'' is the abbreviation of Indo-European.}
% \vspace{-1em}
\end{table}

%% file: table/table2.tex
% distances = ['0.00', '0.06', '0.07', '0.09', '0.09', '0.10', '0.12', '0.12', '0.13', '0.13', '0.13', '0.13', '0.13', '0.14', '0.14', '0.14', '0.14', '0.14', '0.14', '0.15', '0.17', '0.17', '0.18', '0.20', '0.20', '0.23', '0.26', '0.28', '0.33', '0.40', '0.49']
\begin{table*}[t]
	\centering
% 	\resizebox{\linewidth}{!}{%
	\small
	\begin{tabular}{c@{ }||c@{ }|c@{ }|c@{ }||c@{ }|c@{ }|c}
		\hline
		\multirow{2}{*}{Lang} & \multicolumn{3}{c||}{Multilingual Word Embeddings} & \multicolumn{3}{c}{Multilingual BERT} \\
		\cline{2-7} 
		 & (en) & (en-fr) & (en-ru) & (en) & (en-fr) & (en-ru)\\
		\hline
		%=====
            en & 
            \textbf{90.23}/\textbf{88.23} & 90.01/88.08 & 89.93/87.93 & \textbf{93.19}/\textbf{91.21} & 92.81/90.97 & 92.77/90.86 \\
            \hline
            no & 
            80.82/72.94 & 80.60/72.83 & \textbf{80.98}/\textbf{73.10} & \textbf{85.81}/\textbf{79.03} & 85.50/78.64 & 85.43/78.76 \\
            sv & 
            80.33/72.54 & 79.90/72.16 & \textbf{80.43}/\textbf{72.68}
            & 85.61/78.34 & \textbf{85.64}/\textbf{78.58} & 85.44/78.33 \\
            fr & 
            77.71/72.35 & \textbf{78.49}$^\dag$/\textbf{73.30}$^\dag$ & 78.31/73.29 & 85.22/80.78 & 84.76/80.26 & \textbf{85.91}$^\dag$/\textbf{81.63}$^\dag$ \\
            pt & 
            76.41/67.35 & 76.88$^\dag$/67.74 & \textbf{77.09}$^\dag$/\textbf{67.81} & 82.93/73.33 & 82.71/73.13 & \textbf{83.43}$^\dag$/\textbf{73.88}$^\dag$ \\
            da & 
            \textbf{76.58}/\textbf{68.11} & 75.99/67.64 & 76.25/68.03 & 82.36/73.53	& \textbf{82.40}/73.68 & 82.36/\textbf{73.86}$^\dag$ \\
            es & 
            73.76/65.46 & \textbf{74.14}/65.78 & 74.08/\textbf{65.84} & 80.81/72.66	& 81.11/72.80 & \textbf{81.38}$^\dag$/\textbf{73.29}$^\dag$ \\
            it & 
            80.89/75.61 & \textbf{81.33}$^\dag$/\textbf{76.14}$^\dag$ & 80.70/75.57	& 87.07/82.38	& 86.90/82.22 & \textbf{87.41}/\textbf{82.67} \\
            hr & 
            62.21/52.67 & \textbf{63.38}$^\dag$/\textbf{53.83}$^\dag$ & 63.11$^\dag$/53.62$^\dag$ & 72.96/62.65 & 73.39$^\dag$/62.20 & \textbf{74.20}$^\dag$/\textbf{63.55}$^\dag$ \\
            ca & 
            73.18/64.53 & \textbf{73.46}$^\dag$/64.71 & 73.40/\textbf{64.90}$^\dag$ & 80.40/71.42 & 80.30/71.42 & \textbf{80.75}/\textbf{71.78} \\
            pl & 
            74.65/62.72 & 75.65$^\dag$/63.31$^\dag$ & \textbf{75.93}/\textbf{63.60} & 81.51/69.25 & 82.33$^\dag$/69.91$^\dag$ & \textbf{82.48}$^\dag$/\textbf{70.54}$^\dag$ \\
            uk & 
            59.25/51.92 & 60.58$^\dag$/\textbf{52.72}$^\dag$ & \textbf{60.81}$^\dag$/52.66$^\dag$ & 69.98/61.52 & 70.24/61.61 & \textbf{71.21}$^\dag$/\textbf{62.84}$^\dag$ \\
            sl & 
            67.51/56.42	 & 68.14/56.52 & \textbf{68.40}/\textbf{56.87} & 75.15/63.12 & 74.60/62.52 & \textbf{75.50}/\textbf{63.65}$^\dag$ \\
            nl &
            68.54/59.99 & 68.80/60.23 & \textbf{69.23}$^\dag$/\textbf{60.51}$^\dag$ & 76.76/68.35	& \textbf{76.94}/68.28 & 76.89/\textbf{68.76}$^\dag$ \\
            bg & 
            79.09/67.61 & \textbf{80.01}$^\dag$/\textbf{68.42} & 79.72/68.39 & 86.82/75.47 & 87.08/75.40 & \textbf{87.61}$^\dag$/\textbf{75.94}$^\dag$ \\
            ru &
            60.91/52.03 & 61.42$^\dag$/52.27$^\dag$ & \textbf{61.67}$^\dag$/\textbf{52.41}$^\dag$ & 71.92/62.09 & 72.31/62.15 & \textbf{72.88}$^\dag$/\textbf{62.94}$^\dag$ \\
            de & 
            \textbf{71.41/61.97} & 70.70/61.41 & 71.05/61.84 & 78.66/69.81	& 78.04/69.23 & \textbf{79.08}$^\dag$/\textbf{70.26}$^\dag$ \\
            he & 
            55.70/48.08 & \textbf{57.33}$^\dag$/\textbf{49.37}$^\dag$ & 57.15$^\dag$/49.36$^\dag$ & 64.46/\textbf{55.82} & 64.97$^\dag$/55.63 & \textbf{65.30}$^\dag$/55.76 \\
            cs & 
            63.30/54.14 & 63.94$^\dag$/54.63$^\dag$ & \textbf{64.37}$^\dag$/\textbf{55.08}$^\dag$	& 73.78/63.52	& \textbf{74.57}$^\dag$/63.86 & 74.56$^\dag$/\textbf{64.17}$^\dag$ \\
            ro & 
            65.13/53.98	& \textbf{65.86}/\textbf{54.76} & 65.57/54.42 & 75.10/62.99	& 75.85$^\dag$/\textbf{63.92}$^\dag$ & \textbf{76.06}$^\dag$/63.78$^\dag$ \\
            sk & 
            66.79/58.23 & \textbf{67.46}$^\dag$/\textbf{58.77} & 67.42$^\dag$/58.70 & 76.30/67.38	& 77.08$^\dag$/67.57 & \textbf{77.86}$^\dag$/\textbf{68.28}$^\dag$ \\
            id & 
            49.85/44.09 & \textbf{52.05}$^\dag$/\textbf{45.76}$^\dag$ & 51.57/45.31 & 56.80/50.24	& \textbf{57.45}$^\dag$/50.27 & 57.30$^\dag$/\textbf{50.70}$^\dag$ \\
            lv & 
            70.45/49.47 & 70.03/49.38 & \textbf{70.67}$^\dag$/\textbf{49.61}$^\dag$ & \textbf{75.63}/53.93 & 75.27/53.78 & 75.62/\textbf{54.29} \\
            fi & 
            66.11/48.73 & 65.84/48.61 & \textbf{66.28}/\textbf{48.82} & 71.59/\textbf{53.81} & 71.35/53.63 & \textbf{71.74}/53.79 \\
            et & 
            65.01/44.78	& 65.31$^\dag$/45.12$^\dag$ & \textbf{65.38}$^\dag$/\textbf{45.32}$^\dag$ & 71.55/50.98 & \textbf{71.73}/\textbf{51.27} & 71.25/51.16 \\
            ar & 
            37.63/27.48 & 38.72$^\dag$/\textbf{28.00}$^\dag$ & \textbf{38.98}$^\dag$/27.89$^\dag$ & 49.27/37.62 & 50.37$^\dag$/39.37$^\dag$ & \textbf{50.95}$^\dag$/\textbf{39.57}$^\dag$ \\
            la & 
            47.74/34.90 & 48.80$^\dag$/35.64$^\dag$ & \textbf{49.17}$^\dag$/\textbf{35.73}$^\dag$ & 51.83/38.20 & 51.48/38.00 & \textbf{52.20}/\textbf{38.28} \\
            ko & 
            \textbf{34.44}/\textbf{16.18} & 33.98/15.93 & 34.23/16.08 & 38.10/20.62	& 38.03/20.59 & \textbf{38.98}$^\dag$/\textbf{21.54}$^\dag$ \\
            hi & 
            36.34/27.43 & 36.72/27.40 & \textbf{37.37}$^\dag$/\textbf{28.01}$^\dag$ &     45.40/35.03	& \textbf{47.74}$^\dag$/\textbf{35.90}$^\dag$ & 46.10$^\dag$/34.74 \\
            \hline
            Average & 65.92/55.86 & 66.40$^\dag$/56.22$^\dag$ & \textbf{66.53}$^\dag$/\textbf{56.32}$^\dag$ & 73.34/62.93 & 73.55/62.99 & \textbf{73.88}$^\dag$/\textbf{63.43}$^\dag$ \\
		%===========
		\hline
	\end{tabular}
% }
\caption{\label{tab:res_test} Cross-lingual transfer performances (UAS\%/LAS\%, excluding punctuation) of the SelfAtt-Graph parser \cite{ahmad2019} on the test sets. 
In column 1, languages are sorted by the word-ordering distance to English.
(en-fr) and (en-ru) denotes the source-auxiliary language pairs.
`\dag{}' indicates that the adversarially trained model results are statistically significantly better (by permutation test, p $<$ 0.05) than the model trained only on the source language (en).
Results show that the utilization of unlabeled auxiliary language corpora improves cross-lingual transfer performance significantly.
} 
% \vspace{-1em}
\end{table*}

%% file: table/table5.tex
\begin{table}[t]
	\centering
	\resizebox{\linewidth}{!}{%
	\small
	\begin{tabular}{c@{ }|c@{ }|c@{ }|c}
	\hline
	    Lang & (en) & (en-fr) & (en-ru) \\
		\hline
		%=====
            en & 89.65/87.43 & \textbf{89.88}/\textbf{87.66} & 89.67/87.56\\
            \hline
            no & 80.20/72.11 & 80.42/72.49 & \textbf{80.73}$^\dag$/\textbf{72.65}$^\dag$\\
            sv & 81.02/72.95 & 81.14/\textbf{73.44}$^\dag$ & \textbf{81.20}/73.37\\
            fr & 77.42/72.27 & 77.45/72.72 & \textbf{77.78}/\textbf{73.10}\\
            pt & 75.94/67.40 & 76.09/67.47 & \textbf{76.39}$^\dag$/\textbf{67.85}$^\dag$\\
            da & 76.87/68.06 & 77.43$^\dag$/68.62$^\dag$ & \textbf{77.92}$^\dag$/\textbf{69.24}$^\dag$\\
            es & 73.92/65.95 & 74.32$^\dag$/66.35$^\dag$ & \textbf{74.83}$^\dag$/\textbf{66.83}$^\dag$ \\
            it & 80.09/75.36 & 80.98$^\dag$/76.00$^\dag$ & \textbf{81.04}$^\dag$/\textbf{76.06}$^\dag$\\
            hr & 59.53/49.19 & 60.00$^\dag$/50.02$^\dag$ & \textbf{60.16}$^\dag$/\textbf{50.16}$^\dag$\\
            ca & 73.62/64.97 & 73.73/65.11 & \textbf{74.18}$^\dag$/\textbf{65.59}$^\dag$ \\
            pl & 71.48/57.43 &  72.48$^\dag$/\textbf{59.19}$^\dag$ & \textbf{72.55}$^\dag$/58.38$^\dag$\\
            uk & 57.23/49.67 &  58.38$^\dag$/\textbf{51.04}$^\dag$ & \textbf{58.57}$^\dag$/50.88$^\dag$\\
            sl & 65.48/53.40 & 66.11$^\dag$/\textbf{54.21}$^\dag$ & \textbf{66.23}$^\dag$/54.09$^\dag$ \\
            nl & 67.13/59.15 & 67.57/59.71$^\dag$ & \textbf{67.76}$^\dag$/\textbf{59.96}$^\dag$\\
            bg & 77.28/65.77 & 77.79$^\dag$/\textbf{66.66}$^\dag$	& \textbf{78.02}$^\dag$/66.53$^\dag$ \\
            ru & 58.70/49.34  & 59.77$^\dag$/\textbf{50.77}$^\dag$ & \textbf{59.98}$^\dag$/50.51$^\dag$\\
            de & 69.71/58.51 & 70.03/\textbf{59.45}$^\dag$ & \textbf{70.05}/59.38$^\dag$\\
            he & 52.97/45.73 & 53.63$^\dag$/46.49$^\dag$ & \textbf{54.72}$^\dag$/\textbf{47.34}$^\dag$\\
            cs & 60.99/51.63 & 61.60$^\dag$/52.41$^\dag$	& \textbf{61.81}$^\dag$/\textbf{52.45}$^\dag$ \\
            ro & 62.01/51.03 & 62.49/51.30 & \textbf{63.22}$^\dag$/\textbf{51.91}$^\dag$\\
            sk & 64.44/56.01 & 65.03$^\dag$/56.65$^\dag$ & \textbf{65.36}$^\dag$/\textbf{56.67}$^\dag$ \\
            id & 45.08/40.00 & 45.46/40.61$^\dag$ & \textbf{46.82}$^\dag$/\textbf{41.63}$^\dag$\\
            lv & 70.22/48.46 & \textbf{71.08}$^\dag$/\textbf{49.10}$^\dag$ & 70.76/48.86\\
            fi & 65.39/47.78 & \textbf{65.59}/\textbf{48.31}$^\dag$ & 65.42/47.84\\
            et & 64.73/43.84 & 65.01/\textbf{44.27} & \textbf{65.04}/44.16 \\
            ar & 30.98/23.83 & 31.91$^\dag$/24.72$^\dag$ & \textbf{32.83}$^\dag$/\textbf{25.34}$^\dag$\\
            la & \textbf{45.28}/33.08 & 44.94/32.94 & 45.12/\textbf{33.11}\\
            ko & \textbf{33.50}/\textbf{14.36} & 32.87/14.10 & 32.60/14.11\\
            hi & 27.63/19.16 & \textbf{27.66}/\textbf{19.22} & 26.72/18.96\\
            \hline
            Average & 64.09/53.93 & 64.51$^\dag$/54.52$^\dag$ & \textbf{64.74}$^\dag$/\textbf{54.64}$^\dag$ \\
		%===========
		\hline
	\end{tabular}
}
\caption{\label{tab:res_rnnstack} Cross-lingual transfer results (UAS\%/LAS\%, excluding punctuation) of the RNN-Stack parser on the test sets.
`\dag{}' indicates that the adversarially trained model results are statistically significantly better (by permutation test, p $<$ 0.05) than the model trained only on the source language (en).
} 
% \vspace{-1em}
\end{table}

%% file: table/table6.tex
\begin{table*}[t]
\centering
% \resizebox{\linewidth}{!}{%
\small
\begin{tabular}{c|c|c|c|c||c|c|c|c}
\hline
\multirow{3}{*}{AT} & \multicolumn{4}{c||}{SelfAtt-Graph} & \multicolumn{4}{c}{RNN-Stack}\\
\cline{2-5} \cline{6-9}
& \multicolumn{2}{c|}{en-fr} & \multicolumn{2}{c||}{en-ru} & \multicolumn{2}{c|}{en-fr} & \multicolumn{2}{c}{en-ru} \\
\cline{2-3} \cline{4-5} \cline{6-7} \cline{8-9}
& word & sent & word & sent & word & sent & word & sent \\
\hline
GR & 66.19 & 66.21 & 66.38 & 66.38 & \textbf{64.51} & \textbf{64.51} & 64.52 & 64.52 \\
GAN & \textbf{66.40} & 66.29 & \textbf{66.53} & 66.41 & 64.40 & \textbf{64.51} & 64.63 & \textbf{64.74} \\
WGAN & 66.24 & 66.18 & 66.40 & 66.27 & 64.29 & 64.34 & 64.57 & 64.57 \\
\hline
\end{tabular}
% }
\caption{
Average cross-lingual transfer performances (UAS\%, excluding punctuation) on the test sets using different adversarial training objective and setting. % with the best hyper-parameter $\lambda$. 
Multilingual word embeddings are used for these experiments.
}
\label{table:at_abl}
% \vspace{-1em}
\end{table*}

% GR & 66.19/56.05 & 66.38/56.23 & 64.51/54.37 & 64.52/54.36 &  \\
% GAN & \textbf{66.40/56.22} & \textbf{66.53/56.32}  & \textbf{64.51/54.52} & \textbf{64.74/54.64} \\
% WGAN & 66.24/56.08 & 66.40/56.11  & 64.34/54.22 & 64.57/54.40 \\

%% file: table/table4.tex
\begin{table*}[t]
\centering
% \resizebox{\linewidth}{!}{%
\small
\begin{tabular}{c|c|c|c|c||c|c}
\hline
Lang & \multicolumn{2}{c|}{Auxiliary Language Perf.} & \multicolumn{2}{c||}{Average Cross-lingual Perf.} &
\multicolumn{2}{c}{Lang. Test Perf.} \\
% \multirow{2}{*}{AT} & \multirow{2}{*}{MTL} \\ 
\cline{2-3} \cline{4-5} \cline{6-7}
(Src. + Aux.) & AT & MTL & AT & MTL & AT & MTL  \\ 
\hline
en + fr & \textbf{78.49/73.30}$^\dag$ & 78.26/72.98$^\dag$ & \textbf{66.40/56.22} & 66.18/56.04 & \textbf{62.25} & 59.94 \\
en + pt & \textbf{76.53/67.45}$^\dag$ & 75.88/66.75 & \textbf{66.40/56.22} & 66.27/56.08 & 60.17 & \textbf{72.02} \\
en + es & 73.66/65.48 & \textbf{74.04/65.83}$^\dag$ & \textbf{66.38/56.24} & 66.22/56.12 & 56.78 & \textbf{74.52} \\
en + ru & \textbf{61.67/52.41}$^\dag$ & 61.08/52.04 & \textbf{66.53/56.32} & 66.35/56.20 & 37.34 & \textbf{60.56} \\
en + de & \textbf{71.65/62.11}$^\dag$ & 71.17/61.88 & \textbf{66.41/56.13} & 66.18/56.12 & 61.22 & \textbf{72.08} \\
en + la & \textbf{49.22/35.94}$^\dag$ & 48.04/35.09$^\dag$ & \textbf{66.45/56.20} & 66.17/56.05 & 50.04 & \textbf{64.91} \\
\hline
\end{tabular}
% }
\caption{
Comparison between adversarial training (AT) and multi-task learning (MTL) of the contextual encoders.
Columns 2--5 demonstrate the parsing performances (UAS\%/LAS\%, excluding punctuation) on the auxiliary languages and average of the 29 languages.
Columns 6--7 present accuracy (\%) of the language label prediction test.
`\dag{}' indicates that the performance is higher than the baseline performance (shown in the 2nd column of Table \ref{tab:res_test}).
}
% \NoteMA{1. Where are the results of baseline parsers? 2. Shall we add a headline for columns 6-7?}
% \notewa{In this table, we present a comparison between MTL and AT models. I assume by baseline parser, you meant the parser trained only on English language which is shared in prior tables.}
% }
\label{table:at_vs_mtl}
% \vspace{-1em}
\end{table*}

%% file: table/table3.tex
\begin{comment}
\begin{table}[t]
\centering
\resizebox{\linewidth}{!}{%
\small
\begin{tabular}{c|c|c|c}
\hline
Aux. & Avg. Dist. & \multirow{2}{*}{SelfAtt-Graph} & \multirow{2}{*}{RNN-Stack}\\
lang & to other lang & & \\
% Lang & Avg. Dist. & SelfAtt-Graph & RNN-Stack\\
\hline
pt & 0.144 & 66.40/56.22 & 64.41/54.34 \\
ru & 0.146 & \textbf{66.53/56.32} & \textbf{64.74/54.64} \\
de & 0.151 & 66.41/56.13 & 64.57/54.45 \\
es & 0.151 & 66.38/56.24 & 64.52/54.47 \\
fr & 0.160 & 66.40/56.22 & 64.51/54.52 \\
la & 0.242 & 66.45/56.20 & 64.50/54.46 \\
% ru, la & 0.194 & 66.42/56.19 & Running \\
% ru, pt & 0.145 & 66.63/56.34 & Running \\
\hline
\end{tabular}
}
\caption{
Average cross-lingual transfer performances (UAS\%/LAS\%, w/o punctuation) on the test sets to show the effect of different languages playing the role of the auxiliary language during adversarial training.
}
\label{table:lang_comp}
\vspace{-1em}
\end{table}
\end{comment}

\begin{table}[t]
\centering
\resizebox{\linewidth}{!}{%
\small
\begin{tabular}{c|c|c|c}
\hline
Aux. & Avg. Dist. & multilingual & multilingual \\
lang & to other lang & Word Emb. & BERT \\
% Lang & Avg. Dist. & SelfAtt-Graph & RNN-Stack\\
\hline
pt & 0.144 & 66.40/56.22 & 73.47/63.11 \\
ru & 0.146 & \textbf{66.53/56.32} & 73.88/63.43 \\
de & 0.151 & 66.41/56.13 & \textbf{73.92/63.56} \\
es & 0.151 & 66.38/56.24 & 71.71/62.49 \\
fr & 0.160 & 66.40/56.22 & 73.55/62.99 \\
la & 0.242 & 66.45/56.20 & 73.69/63.29 \\
% ru, la & 0.194 & 66.42/56.19 & Running \\
% ru, pt & 0.145 & 66.63/56.34 & Running \\
\hline
\end{tabular}
}
\caption{
Average cross-lingual transfer performances (UAS\%/LAS\%, w/o punctuation) on the test sets using SelfAtt-Graph parser when different languages play the role of the auxiliary language in adversarial training.
}
\label{table:lang_comp}
% \vspace{-1em}
\end{table}

%% file: table/table7.tex
\begin{table}[t]
	\centering
	\resizebox{\linewidth}{!}{%
	\small
	\begin{tabular}{@{ }c@{ }|@{ }c@{ }|@{ }c@{ }|@{ }c@{ }|@{ }c@{ }}
		\hline
		Lang & (en,ru) - en & (en,fr) - en & (en,de) - en & (en,la) - en \\
		\hline
        \multicolumn{5}{c}{IE.Slavic Family} \\
		\hline
        hr	& 1.24/0.90 & 0.43/-0.45 & 1.52/1.02 & 0.06/-0.13 \\
        sl	& 0.35/0.53	& -0.55/-0.60 &	-0.04/0.14 & -0.17/-0.50 \\
        uk	& 1.23/1.32	& 0.26/0.09	& 1.54/1.33	& -0.29/-0.09 \\
        pl	& 0.97/1.29	& 0.82/0.66 & 0.82/0.98 & 1.03/0.98 \\
        bg	& 0.79/0.47	& 0.26/-0.07 &	0.49/0.41 &	0.01/0.04 \\
        ru	& 0.96/0.85	& 0.39/0.06	 & 1.07/1.11 & 0.20/0.34 \\
        cs	& 0.78/0.65	& 0.79/0.34 & 0.91/0.81	& -0.08/0.05 \\
        sk  & 1.56/0.90 & 0.78/0.19 & 1.88/1.04 & 0.56/0.66 \\
        \hline
        \textbf{Avg.} & \textbf{0.98/0.86} & \textbf{0.4/0.03} & \textbf{1.02/0.86} & \textbf{0.17/0.17} \\
		\hline
        \multicolumn{5}{c}{IE.Romance Family} \\
		\hline
		pt	& 0.50/0.55	& -0.22/-0.20 & 0.54/0.80 &	0.49/0.60 \\
        fr	& 0.69/0.85	& -0.46/-0.52 & 0.49/0.16 & 0.95/0.86 \\
        es	& 0.57/0.63	& 0.30/0.14	& 0.45/0.39	& 0.44/0.51 \\
        it  & 0.34/0.29	& -0.17/-0.16 &	-0.22/-0.17 & 0.26/0.18 \\
        ca  & 0.35/0.36	& -0.10/0.00 & 0.64/0.70 & 0.10/0.28 \\
        ro  & 0.96/0.79	& 0.75/0.93	& 1.32/1.32	& 1.62/1.73 \\
        \hline
        \textbf{Avg.} & \textbf{0.57/0.58} & \textbf{0.02/0.03} & \textbf{0.54/0.53} & \textbf{0.64/0.69} \\
        \hline
        \multicolumn{5}{c}{IE.Germanic Family} \\
		\hline
		%=====
		en	& -0.42/-0.35 & -0.38/-0.24 & -0.35/-0.25 &	-0.15/-0.20 \\
        no	& -0.38/-0.27 & -0.31/-0.39 & -0.41/-0.15 & -0.22/-0.24 \\
        sv	& -0.17/-0.01 & 0.03/0.24 & -0.12/0.35 & -0.02/0.18 \\
        da	& 0.00/0.33 & 0.04/0.15 & -0.15/0.08 & -0.46/-0.25 \\
        nl	& 0.13/0.41 & 0.18/-0.07 & 0.95/0.89 & 0.57/0.42 \\
        de	& 0.42/0.45 & -0.62/-0.58 & 1.41/1.40 & 0.25/0.43 \\
        \hline
        \textbf{Avg.} & \textbf{-0.07/0.09} & \textbf{-0.18/-0.15} & \textbf{0.22/0.39} & \textbf{0.00/0.06} \\
		%===========
		\hline
	\end{tabular}
}
\caption{
\label{tab:lang_family_test} 
Average cross-lingual performance difference between the SelfAtt-Graph parser trained on the source (en) and an auxiliary (x) language and the SelfAtt-Graph parser trained only on English (en) language (UAS\%/LAS\%, excluding punctuation).
We use multilingual BERT in this set of experiments.
% $^1$, $^2$, $^3$ indicates the IE.Germanic, IE.Romance, and IE.Slavic languages respectively.
% `\dag{}' means that the best transfer model is statistically significantly better (by paired bootstrap test, p $<$ 0.05) than all other transfer models.
} 
% \vspace{-1em}
\end{table}

%% file: relwork.tex
\section{Related Work}

\paragraph{Unsupervised Cross-lingual Parsing.}
Unsupervised cross-lingual transfer for dependency parsing has been studied over the past few years \cite{agic2014cross, ma2014unsupervised, xiao2014distributed, tiedemann2015cross, guo2015cross, aufrant2015zero, rasooli2015density, duong-etal-2015-cross, schlichtkrull2017cross, ahmad2019, rasooli2019low, he2019cross}.
% Here, by ``unsupervised transfer'', we refer to the setting where there are no annotated target instances. 
% That is, a parsing model trained only on the source language is directly transferred to the target languages.
Here, ``unsupervised transfer'' refers to the setting where a parsing model trained only on the source language is directly transferred to the target languages. 
In this work, we relax the setting by allowing unlabeled data from one or more auxiliary (helper) languages other than the source language. 
This setting has been explored in a few prior works. 
\citet{cohen-etal-2011-unsupervised} learn a generative target language parser with unannotated target data as a linear interpolation of the source language parsers. 
\citet{tackstrom2013target} adopt unlabeled target language data and a learning method that can incorporate diverse knowledge sources through ambiguous labeling for transfer parsing.
In comparison, we leverage unlabeled auxiliary language data to learn language-agnostic contextual representations to improve cross-lingual transfer.
% in a wide spectrum of languages.

\paragraph{Multilingual Representation Learning.}
The basic of the unsupervised cross-lingual parsing is that we can align the representations of different languages into the same space, at least at the word level.
The recent development of bilingual or multilingual word embeddings provide us with such shared representations. 
We refer the readers to the surveys of \citet{ruder2017survey} and \citet{glavas2019} for details. 
% The main idea is that since word embeddings from different languages are aligned in the same space after we train a model with the source language embeddings, we can directly replace the source embeddings with the target ones for the transfer. 
The main idea is that we can train a model on top of the source language embeddings which are aligned to the same space as the target language embeddings and thus all the model parameters can be directly shared across languages.
During transfer to a target language, we simply replace the source language embeddings with the target language embeddings.
% Since the word embeddings from different languages are aligned in the same space, we can directly replace the source language embeddings with the target language embeddings while transfer.
This idea is further extended to learn multilingual contextualized word representations, for example, multilingual BERT \cite{devlin2018bert}, have been shown very effective for many cross-lingual transfer tasks \cite{wu2019beto}.
In this work, we show that further improvements can be achieved by adaptating the contextual encoders via unlabeled auxiliary languages even when the encoders are trained on top of multilingual BERT.

\paragraph{Adversarial Training.}
The concept of adversarial training via Generative Adversarial Networks (GANs) \cite{goodfellow2014generative, szegedy2013intriguing, goodfellow2014explaining} was initially introduced in computer vision for image classification and received enormous success in improving model's robustness on input images with perturbations.
% GANs are well known for suffering from training instability and to improve many variants are proposed in prior works \cite{arjovsky2017wasserstein, gulrajani2017improved}.
Later many variants of GANs \cite{arjovsky2017wasserstein, gulrajani2017improved} were proposed to improve its' training stability. 
% \notewa{Not sure if I said too much here.}
In NLP, adversarial training was first utilized for domain adaptation \cite{ganin2016domain}.
Since then adversarial training has started to receive an increasing interest in the NLP community and applied to many NLP applications  including part-of-speech (POS) tagging \cite{gui2017part, yasunaga-etal-2018-robust}, dependency parsing \cite{sato2017adversarial}, relation extraction \cite{wu2017adversarial}, text classification \cite{miyato2016adversarial, liu-etal-2017-adversarial, chen2018multinomial}, dialogue generation \cite{li-etal-2017-adversarial}.

% Adversarial training for cross-lingual NLP applications has been explored in a few recent studies \cite{joty2017cross, adel2018adversarial, wang2018transition, chen2018adversarial, zou2018adversarial}.

In the context of cross-lingual NLP tasks, many recent works adopted adversarial training, such as in sequence tagging \cite{adel2018adversarial}, text classification \cite{xu-yang-2017-cross, chen2018adversarial}, word embedding induction \cite{zhang2017adversarial, lample2018word}, relation classification \cite{zou2018adversarial}, opinion mining \cite{wang2018transition}, and question-question similarity reranking \cite{joty2017cross}.
However, existing approaches only consider using the \emph{target} language as the auxiliary language.
It is unclear whether the language invariant representations learned by previously proposed methods can perform well on a wide variety of \emph{unseen} languages.
To the best of our knowledge, we are the first to study the effects of language-agnostic representations on a broad spectrum of languages.
% Although all these previous works exploit the use case of adversarial training, learning language agnostic representations, the usefulness of such representations is not tested in a broad spectrum of languages.

%% file: conclusion.tex
\section{Conclusion}
In this paper, we study learning language invariant contextual encoders for cross-lingual transfer. Specifically, we leverage unlabeled sentences from auxiliary languages and adversarial training to induce language-agnostic encoders to improve the performances of the cross-lingual dependency parsing.
% We couple dependency parsing task with an auxiliary language identification task where a contextual encoder serves both the tasks.
% The contextual encoder learns to encode informative features by collaborating with a decoder that predicts the dependency structure of the input sequences while acts adversarially with a language identity classifier to capture language invariant features.
Experiments and analysis using English as the source language and six foreign languages as the auxiliary languages not only show improvements on cross-lingual dependency parsing, but also demonstrates that contextual encoders successfully learns not to capture language-dependent features through adversarial training.
In the future, we plan to investigate the effectiveness of adversarial training for multi-source transfer to parsing and other cross-lingual NLP applications.

% In this work, we train contextual encoders to learn one representation per word and apply adversarial training to make them language-agnostic.
% In our future work, we want to train contextual encoders to learn separate representations that are language-agnostic and language-dependent and apply adversarial training to disentangle them.
% In this way, the resulting language-agnostic representations would perform better in cross-lingual transfer settings. 
% Moreover, we want to investigate the advantages of adversarial training for multi-source transfer parsing.

% This work opens up the door for further research to study language adaptation to contextual encoders.
% For example, most of the contextual encoders employed in deep neural NLP models are consists of multiple layers where the lower and upper layers learn to encode syntactic and semantic information, respectively.
% To make the encoders language agnostic, adversarial training may be applied to each layer in a controlled way to purge language-dependent features.
% Besides, we motivate the need for a better evaluation technique to evaluate whether an encoder learns language-agnostic or dependent features because looking at the targeted task results may not be conclusive.